\documentclass{talan_simple}
\usepackage{graphicx}
\usepackage{amsfonts}
\usepackage{amssymb}
\usepackage{colortbl}
\usepackage{amsmath}
\usepackage{booktabs}
\usepackage{url}
\usepackage[T1]{fontenc}
\begin{document}
\newcommand{\keywords}[1]{%
  \par\noindent\textbf{Keywords:} \begingroup\renewcommand{\and}{, }#1\endgroup%
}

\title{\raisebox{-0.2\height}{%
  \includegraphics[height=1.1em]{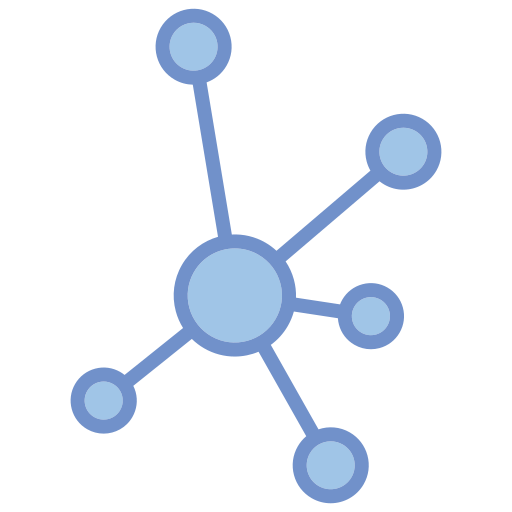}%
}\hspace{0.02em}\Large TopoLoRA-SAM: Topology-Aware Parameter-Efficient Adaptation of Foundation Segmenters
for Thin-Structure and Cross-Domain Binary Semantic Segmentation}

%
\author[1]{Salim Khazem}
\affiliation[1]{Talan Research \& Innovation Center, Paris, France}
\abstract{%
\\ 
Foundation segmentation models such as the Segment Anything Model (SAM) exhibit strong zero-shot generalization through large-scale pretraining, but adapting them to domain-specific semantic segmentation remains challenging, particularly for thin structures (e.g., retinal vessels) and noisy modalities (e.g., SAR imagery). Full fine-tuning is computationally expensive and risks catastrophic forgetting. We propose \textbf{TopoLoRA-SAM}, a topology-aware and parameter-efficient adaptation framework for binary semantic segmentation. TopoLoRA-SAM injects Low-Rank Adaptation (LoRA) into the frozen ViT encoder, augmented with a lightweight spatial convolutional adapter and optional topology-aware supervision via differentiable clDice. We evaluate our approach on five benchmarks spanning retinal vessel segmentation (DRIVE, STARE, CHASE\_DB1), polyp segmentation (Kvasir-SEG), and SAR sea/land segmentation (SL-SSDD), comparing against U-Net, DeepLabV3+, SegFormer, and Mask2Former. TopoLoRA-SAM achieves the best retina-average Dice and the best overall average Dice across datasets, while training only \textbf{5.2\%} of model parameters ($\sim$4.9M). On the challenging CHASE\_DB1 dataset, our method substantially improves segmentation accuracy and robustness, demonstrating that topology-aware parameter-efficient adaptation can match or exceed fully fine-tuned specialist models. Code is available at : \url{https://github.com/salimkhazem/Seglab.git}
}

\date{\today}

\correspondence{\email{salim.khazem@talan.com}}

\codeurl{\url{https://github.com/salimkhazem/Seglab.git}}

\metadata[ORCID]{0000-0001-5958-6120}
\maketitle              

\begin{figure}[t]
  \centering
  \includegraphics[width=\linewidth]{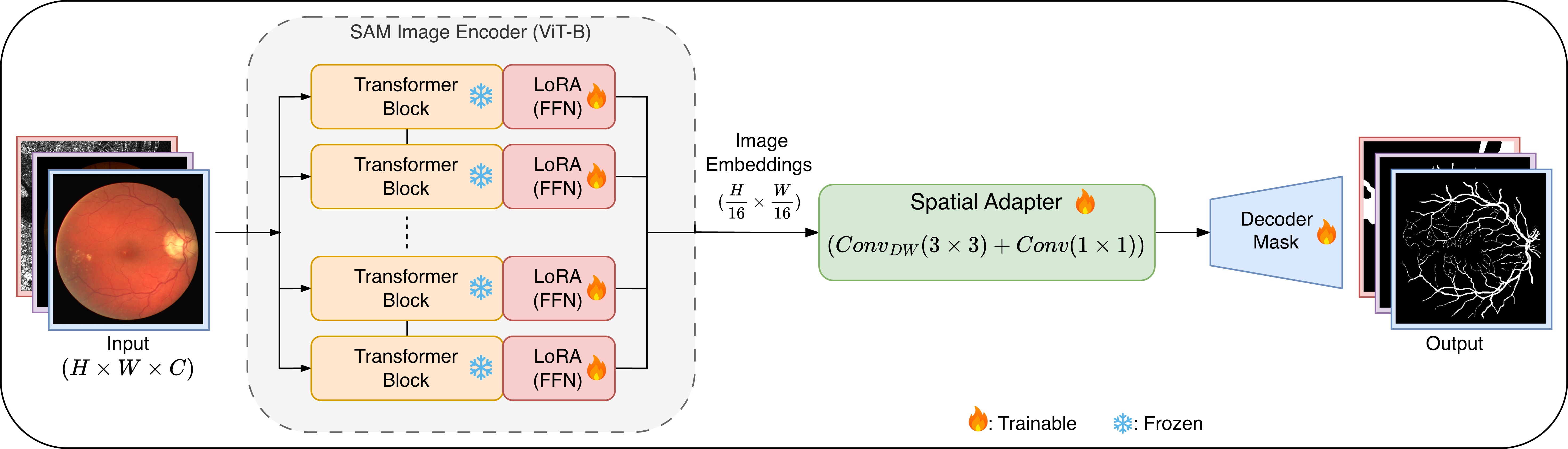} 
  \caption{\textbf{TopoLoRA-SAM architecture overview.} We freeze the SAM ViT-B image encoder and inject LoRA modules (red) into the feed-forward network (FFN) layers of each transformer block. A lightweight depthwise-separable convolutional adapter (green) refines the high-resolution embedding tensor before mask decoding. Training uses a topology-aware loss combining BCE, Dice, and clDice.}
  \label{fig:method}
\end{figure}

\section{Introduction}
Semantic segmentation underpins numerous critical applications in medical imaging and remote sensing, from automated diabetic retinopathy screening~\cite{staal2004drive} to maritime surveillance and environmental monitoring~\cite{slssdd2020}. While recent foundation models such as the Segment Anything Model 
(SAM)~\cite{kirillov2023segment} have achieved remarkable zero-shot generalization through web-scale pretraining, significant challenges remain when adapting these models to specialized domains with distinct visual characteristics. Two challenges are particularly acute in this setting. First, \textbf{thin and elongated structures} such as retinal vasculature or road networks exhibit extreme aspect ratios, where small pixel-level errors can disconnect entire branches; standard region-based losses (e.g., Dice, cross-entropy) are largely insensitive to such topological violations~\cite{shit2021cldice,hu2019topoloss}. Second, \textbf{cross-domain visual shift} arises in modalities such as SAR or fundoscopy, whose texture, noise characteristics, and semantics differ substantially from natural images, limiting the transferability of web-pretrained representations~\cite{chen2022domain}.

Full fine-tuning of billion-parameter foundation models is computationally prohibitive and risks catastrophic forgetting of generalizable representations~\cite{hu2022lora}. Parameter-efficient fine-tuning (PEFT) methods such as LoRA~\cite{hu2022lora}, adapters~\cite{houlsby2019adapters,chen2022adaptformer},
and visual prompt tuning~\cite{jia2022vpt} offer compelling alternatives by updating only a small fraction of parameters while preserving pretrained knowledge. Recent work has begun exploring PEFT for SAM adaptation~\cite{chen2023samadapter,ma2024medsam,zhang2023customsam}, yet systematic studies investigating
\emph{topology-aware} adaptation for domain-specific binary segmentation remain limited. 

In this work, we introduce \textbf{TopoLoRA-SAM}, a principled framework that unifies parameter-efficient adaptation with topology-preserving supervision for binary semantic segmentation across diverse domains. Our approach freezes the SAM ViT-B image encoder and injects trainable LoRA modules into its feed-forward layers, complemented by a lightweight depthwise-separable convolutional adapter operating on high-resolution feature maps. This design enables fine-grained adaptation while preserving the generalizable representations learned during pretraining. To address topological integrity critical for thin structures we incorporate the differentiable clDice loss~\cite{shit2021cldice}, which explicitly optimizes skeleton overlap and encourages connected centerlines. Our contributions are: (i) a topology-aware, parameter-efficient adaptation of SAM
combining LoRA and a lightweight spatial adapter; (ii) a benchmark across five binary segmentation datasets (retinal vessels, polyps, SAR sea/land) with region, boundary, topology, and calibration metrics; and (iii) strong performance on CHASE\_DB1 and best retina-average Dice while training only 5.2\% of model parameters, with a fully reproducible codebase. In this work, ideas related to simplified geometric representations for efficient visual learning~\cite{khazem2025polygonet} and lightweight adaptation via visual prompting~\cite{khazem2025multi} are leveraged within a foundation-model segmentation framework.

  
  
  

\section{Related Work}
\paragraph{\textbf{Semantic segmentation architectures.}}
Fully convolutional networks revolutionized semantic segmentation, with U-Net~\cite{ronneberger2015unet} establishing the encoder-decoder paradigm with skip connections that remains foundational in medical imaging. DeepLabV3+~\cite{chen2018deeplabv3plus} introduced atrous spatial pyramid pooling for multi-scale context aggregation, while recent self-configuring approaches like nnU-Net~\cite{isensee2021nnunet}
achieve strong performance through systematic hyperparameter optimization. ~\cite{khazem2023deep, khazem2023improving} used these kind of architectures for different applications. Transformer-based architectures have emerged as powerful alternatives: SegFormer~\cite{xie2021segformer} proposes an efficient hierarchical design with lightweight MLP decoders, while Mask2Former~\cite{cheng2022mask2former} unifies instance, semantic, and panoptic segmentation through masked attention. These architectures build upon foundational work on vision transformers~\cite{dosovitskiy2021vit} and hierarchical designs~\cite{liu2021swin}.

\paragraph{\textbf{Foundation models for segmentation.}}
The Segment Anything Model~\cite{kirillov2023segment} represents a paradigm shift, trained on over 1 billion masks to enable promptable zero-shot segmentation. While SAM excels at interactive mask generation, adapting it to \emph{semantic} segmentation particularly for domain-specific binary masks without
prompts requires careful consideration of the prompt-mask relationship. SAM 2~\cite{ravi2024sam2} extends this capability to video, demonstrating the scalability of the prompting paradigm. However, for domains with significant distribution shift (medical imaging, remote sensing), prompt-free adaptation with task-specific tuning remains essential~\cite{khazem2025byo}.

\paragraph{\textbf{SAM adaptation for specialized domains.}}
Several concurrent works address SAM adaptation for medical imaging. MedSAM~\cite{ma2024medsam} fine-tunes SAM on large-scale medical datasets, demonstrating strong generalization across modalities. SAM-Adapter~\cite{chen2023samadapter} proposes domain-specific adapter modules for underperformed scenes, while
SAM-Med2D~\cite{cheng2023sammed2d} focuses on 2D medical image segmentation. HQ-SAM~\cite{ke2024hqsam} introduces a high-quality output token for improved mask boundaries. PerSAM~\cite{zhang2023persam} enables one-shot personalization through training-free prompt learning. Our work differs by jointly addressing
parameter efficiency (LoRA + adapters) and topology preservation (clDice) for thin-structure domains where connectivity is critical.

\paragraph{\textbf{Parameter-efficient fine-tuning.}}
PEFT methods enable foundation model adaptation with minimal parameter overhead. LoRA~\cite{hu2022lora} reparameterizes weight updates as low-rank matrices, achieving comparable performance to full fine-tuning on language models. AdaLoRA~\cite{zhang2023adalora} extends this with adaptive rank allocation, while DoRA~\cite{liu2024dora} decomposes weights into magnitude and direction for improved learning dynamics. Adapter modules~\cite{houlsby2019adapters} insert trainable bottleneck layers between transformer blocks, and
AdaptFormer~\cite{chen2022adaptformer} adapts this for vision. Visual Prompt Tuning (VPT)~\cite{jia2022vpt} prepends learnable tokens to input sequences. Scale-and-shift tuning (SSF)~\cite{lian2022scaling} offers an even more parameter-efficient alternative. These methods share the goal of preserving pretrained knowledge while enabling task-specific adaptation. 

\paragraph{\textbf{Topology-aware losses and metrics.}}
Standard overlap metrics (Dice~\cite{milletari2016vnet}, IoU) are insensitive to topological errors that break connectivity in thin structures. clDice~\cite{shit2021cldice} addresses this by computing skeleton overlap using differentiable soft-skeletonization, encouraging connected centerlines in tubular structures. Complementary approaches include TopoLoss~\cite{hu2019topoloss}, which uses persistent homology to
penalize Betti number mismatches, and Betti matching~\cite{stucki2023bettimatching}, which aligns persistence barcodes for topologically faithful predictions. Boundary focused metrics such as BFScore~\cite{csurka2013bfscore} evaluate contour precision. For probabilistic outputs, calibration metrics like Expected Calibration Error (ECE)~\cite{guo2017calibration,naeini2015ece} assess reliability, which is
critical for clinical deployment~\cite{mehrtash2020confidence}. Recent work emphasizes the importance of comprehensive metric evaluation~\cite{reinke2024metrics}.


\subsection{Problem Formulation}
\label{sec:problem}

Given an input image $\mathbf{x} \in \mathbb{R}^{H \times W \times 3}$, we aim to predict a binary semantic mask $\hat{\mathbf{y}} \in [0,1]^{H \times W}$ indicating foreground probability (vessel, polyp, or sea regions). Unlike SAM's promptable paradigm that produces instance-agnostic masks conditioned on
user-provided points or boxes, our task requires \emph{semantic} predictions without runtime prompts. This necessitates adapting SAM's encoder-decoder architecture to directly output semantic masks through end-to-end training.

\subsection{SAM Architecture and Prompt-Free Decoding}
\label{sec:sam-backbone}

We adopt the SAM ViT-B architecture~\cite{kirillov2023segment}, comprising:
(i) a Vision Transformer (ViT) image encoder with 12 transformer blocks,
producing embeddings $\mathbf{z} \in \mathbb{R}^{256 \times \frac{H}{16} \times \frac{W}{16}}$;
(ii) a prompt encoder mapping user interactions to embedding space; and
(iii) a lightweight mask decoder that combines image and prompt embeddings
via cross-attention. For semantic segmentation, we operate in \textbf{prompt-free
mode} by providing null prompt embeddings ($\varnothing$), treating the mask
decoder as a semantic prediction head. The decoder outputs a single binary
mask, upsampled to the target resolution via bilinear interpolation.

\subsection{Low-Rank Adaptation (LoRA)}
\label{sec:lora}

To enable efficient adaptation while preserving pretrained representations, we
inject LoRA modules~\cite{hu2022lora} into the frozen image encoder. For a
pretrained linear layer with weight matrix $\mathbf{W}_0 \in \mathbb{R}^{d_{\text{out}} \times d_{\text{in}}}$,
LoRA reparameterizes the forward pass as:
\begin{equation}
\mathbf{h} = \mathbf{W}_0 \mathbf{x} + \Delta\mathbf{W} \mathbf{x} = \mathbf{W}_0 \mathbf{x} + \mathbf{B}\mathbf{A}\mathbf{x},
\end{equation}
where $\mathbf{A} \in \mathbb{R}^{r \times d_{\text{in}}}$ and $\mathbf{B} \in \mathbb{R}^{d_{\text{out}} \times r}$
are trainable low-rank factors with rank $r \ll \min(d_{\text{in}}, d_{\text{out}})$.
We initialize $\mathbf{A}$ with Kaiming uniform and $\mathbf{B}$ with zeros,
ensuring $\Delta\mathbf{W} = \mathbf{0}$ at initialization for stable training.
A scaling factor $\alpha/r$ modulates the magnitude of low-rank updates.

We target the \textbf{feed-forward network (FFN)} layers within each transformer
block, specifically the \texttt{mlp.lin1} and \texttt{mlp.lin2} projections.
This design choice follows the intuition that FFN layers capture task-specific
feature transformations, while attention weights encode more transferable
relational patterns. With $r=16$ and LoRA applied to all 12 blocks, we add
approximately 2.4M trainable parameters to the encoder.

\subsection{Spatial Convolutional Adapter}
\label{sec:adapter}

While LoRA enables semantic adaptation of the encoder's representational
capacity, thin structures require fine-grained spatial reasoning at higher
resolutions than SAM's $16\times$ downsampled embeddings. We introduce a
lightweight \textbf{depthwise-separable convolutional adapter} applied to the
image embedding tensor $\mathbf{z}$:
\begin{equation}
\mathbf{z}' = \mathbf{z} + \text{Conv}_{1\times1}\left(\text{ReLU}\left(\text{DepthwiseConv}_{3\times3}(\mathbf{z})\right)\right).
\end{equation}
The adapter comprises a $3\times3$ depthwise convolution (256 groups) followed
by ReLU activation and $1\times1$ pointwise projection, wrapped in a residual
connection. This adds only $\sim$66K parameters while enabling local spatial
refinement crucial for thin-structure boundary preservation.

\subsection{Topology-Aware Training Objective}
\label{sec:loss}

Our training objective combines multiple complementary loss terms:
\begin{equation}
\mathcal{L} = \lambda_{\text{bce}} \mathcal{L}_{\text{BCE}} + \lambda_{\text{dice}} \mathcal{L}_{\text{Dice}} + \lambda_{\text{cl}} \mathcal{L}_{\text{clDice}} + \lambda_{\text{bd}} \mathcal{L}_{\text{Boundary}},
\label{eq:loss}
\end{equation}
with default weights $\lambda_{\text{bce}} = 1.0$, $\lambda_{\text{dice}} = 1.0$,
$\lambda_{\text{cl}} = 0.5$, and $\lambda_{\text{bd}} = 0.0$.

\paragraph{Binary cross-entropy (BCE)} provides pixel-wise gradient signal:
\begin{equation}
\mathcal{L}_{\text{BCE}} = -\frac{1}{N}\sum_{i=1}^{N}\left[y_i \log(\hat{y}_i) + (1-y_i)\log(1-\hat{y}_i)\right].
\end{equation}

\paragraph{Soft Dice loss} optimizes region overlap:
\begin{equation}
\mathcal{L}_{\text{Dice}} = 1 - \frac{2\sum_i y_i \hat{y}_i + \epsilon}{\sum_i y_i + \sum_i \hat{y}_i + \epsilon}.
\end{equation}

\paragraph{Centerline Dice (clDice)~\cite{shit2021cldice}} enforces topological
connectivity by computing Dice on soft-skeletonized predictions:
\begin{equation}
\mathcal{L}_{\text{clDice}} = 1 - 2 \cdot \frac{\text{Tprec}(\hat{y}, y) \cdot \text{Tsens}(\hat{y}, y)}{\text{Tprec}(\hat{y}, y) + \text{Tsens}(\hat{y}, y)},
\end{equation}
where $\text{Tprec}$ measures how much of the predicted skeleton overlaps with
the ground-truth mask, and $\text{Tsens}$ measures coverage of the GT skeleton.
Soft skeletonization is implemented via iterated morphological operations
(min-pooling) applied to the soft prediction map.

\begin{table}[t]
  \centering
  \caption{\textbf{Parameter efficiency comparison.} TopoLoRA-SAM trains only
  5.2\% of parameters while achieving competitive or superior performance.}
  \label{tab:params}
  \small
  \begin{tabular}{lrrr}
  \toprule
  \textbf{Model} & \textbf{Total (M)} & \textbf{Trainable (M)} & \textbf{Trainable \%} \\
  \midrule
  U-Net (ResNet34) & 24.4 & 24.4 & 100.0\% \\
  DeepLabV3+ (ResNet50) & 39.8 & 39.8 & 100.0\% \\
  SegFormer (MiT-B0) & 3.7 & 3.7 & 100.0\% \\
  Mask2Former (Swin-T) & 47.4 & 47.4 & 100.0\% \\
  \midrule
  \textbf{TopoLoRA-SAM (ours)} & 93.7 & 4.9 & \textbf{5.2\%} \\
  \bottomrule
  \end{tabular}
\end{table}

\subsection{Implementation Details}
\label{sec:implementation}

We use SAM ViT-B pretrained weights and freeze all encoder parameters except LoRA modules. The mask decoder remains trainable ($\sim$2.4M parameters). Total trainable parameters: $\sim$4.9M (LoRA: 2.4M, adapter: 66K, decoder: 2.4M), compared to 93.7M total. Table~\ref{tab:params} summarizes parameter efficiency. Training uses AdamW with learning rate $10^{-4}$, cosine decay, and 50 epochs.

\section{Experimental Setup} 
\subsection{Datasets}
\label{sec:datasets}

We evaluate on five binary segmentation benchmarks spanning three distinct application domains:

\paragraph{\textbf{Retinal vessel segmentation.}} DRIVE~\cite{staal2004drive} contains 40 color fundus images (584$\times$565) with manual vessel annotations from expert ophthalmologists, split into 20 train
and 20 test images. STARE~\cite{hoover2000stare} provides 20 images (700$\times$605) with vessels and pathological abnormalities. CHASE\_DB1~\cite{fraz2012chase} comprises 28 high-resolution images (999$\times$960) from a pediatric population study. These datasets exemplify thin-structure segmentation where topological connectivity is clinically significant for vascular analysis.

\paragraph{\textbf{Polyp segmentation.}}
Kvasir-SEG~\cite{jha2020kvasirseg} contains 1,000 colonoscopy images with corresponding polyp masks, representing an important task for early colorectal cancer detection. Images vary in resolution and polyp appearance.

\paragraph{\textbf{SAR sea/land segmentation.}}
SL-SSDD~\cite{slssdd2020} provides synthetic aperture radar (SAR) imagery for maritime boundary delineation, challenging due to speckle noise, low contrast, and significant distribution shift from natural images.

\subsection{Preprocessing and Augmentations}
\label{sec:preprocessing}

All datasets are standardized to RGB format with ImageNet normalization
($\mu=[0.485, 0.456, 0.406]$, $\sigma=[0.229, 0.224, 0.225]$). We use dataset-specific
resolutions: 384$\times$384 for retinal datasets, 512$\times$512 for Kvasir-SEG
and SL-SSDD. Training augmentations include random horizontal/vertical flips,
random scaling (0.5--1.5$\times$), and center crops with reflection padding.
Color jitter is disabled for SAR imagery to preserve modality-specific characteristics.

\subsection{Baseline Architectures}
\label{sec:baselines}
We compare five segmentation model families spanning convolutional, transformer-based, and foundation-model paradigms. As convolutional baselines, we use U-Net~\cite{ronneberger2015unet} with a ResNet34~\cite{he2016resnet} encoder and DeepLabV3+~\cite{chen2018deeplabv3plus} with a ResNet50 encoder. For
transformer-based segmentation, we evaluate SegFormer (MiT-B0)~\cite{xie2021segformer} and Mask2Former~\cite{cheng2022mask2former} with a Swin-T backbone~\cite{liu2021swin}. Finally, we include our TopoLoRA-SAM model, which adapts SAM ViT-B~\cite{kirillov2023segment} using Low-Rank Adaptation, a lightweight spatial adapter, and optional topology-aware clDice regularization. All models use ImageNet-pretrained backbones (or SAM-pretrained weights) and are trained with an identical BCE+Dice objective to ensure fair comparison.

\subsection{Training Configuration}
\label{sec:training}
All experiments follow a unified training protocol. We use the AdamW optimizer with $\beta_1=0.9$, $\beta_2=0.999$, and a weight decay of $10^{-4}$. The learning rate is set to $10^{-4}$ and scheduled using cosine annealing down to a minimum value of $10^{-6}$. All models are trained for 50 epochs. A batch size of 4 is used for all baseline models, while SAM-based models are trained with a batch size of 1 and gradient accumulation over 4 steps to maintain a comparable effective batch size. Training is performed using mixed-precision (FP16) via PyTorch Automatic Mixed Precision (AMP). For robustness, all reported results are averaged over three random seeds (0, 1, and 2), and we report the mean and standard deviation.







\subsection{Evaluation Metrics}
\label{sec:metrics}

We use a comprehensive evaluation suite covering region overlap, boundary accuracy, topology preservation, and calibration. Region-level performance is measured using the Dice coefficient (F1-score) and Intersection-over-Union(IoU), with precision and recall reported to characterize false positive and false negative behavior. Boundary quality is evaluated using the BFScore~\cite{csurka2013bfscore}, which computes an F1-score on contour pixels within a tolerance margin. To assess topology preservation for thin structures,
we report centerline Dice (clDice)~\cite{shit2021cldice} on retinal datasets, capturing connectivity via skeleton overlap. Finally, probabilistic calibration is measured using Expected Calibration Error (ECE)~\cite{guo2017calibration}, which quantifies the alignment between prediction confidence and empirical
accuracy; lower values indicate better-calibrated models, an important consideration for clinical decision support.
\subsection{Computational Resources}
\label{sec:compute}

All experiments run on a single NVIDIA RTX A6000 ada (50GB) GPU per training job, with multi-GPU parallelism used only to run independent experiments concurrently. The complete benchmark (5 datasets $\times$ 5 models $\times$ 3 seeds + ablations + cross-dataset) requires approximately \textbf{29.7 GPU-hours}, dominated by
Mask2Former and SAM runs due to larger memory footprint and longer convergence.

\section{Results}

\begin{table}[t]
  \centering
  \caption{\textbf{Benchmark Dice scores} (mean$\pm$std across 3 seeds). Bold
  indicates best performance per dataset. TopoLoRA-SAM achieves the best overall
  average and best retina-average Dice while training only 5.2\% of parameters.}
  \label{tab:main-dice}
  \small
  \resizebox{\textwidth}{!}{\begin{tabular}{llllll}
\toprule
 & U-Net (R34) & DeepLabV3+ (R50) & SegFormer (B0) & Mask2Former (Swin-T) & TopoLoRA-SAM (ours) \\
\midrule
DRIVE & 0.590±0.059 & 0.589±0.027 & 0.541±0.041 & 0.689±0.011 & \textbf{0.690±0.018} \\
STARE & 0.433±0.131 & 0.478±0.056 & 0.374±0.016 & \textbf{0.644±0.023} & 0.565±0.048 \\
CHASE\_DB1 & 0.412±0.050 & 0.345±0.123 & 0.424±0.011 & 0.285±0.293 & \textbf{0.569±0.016} \\
Kvasir-SEG & 0.922±0.004 & 0.917±0.003 & 0.920±0.005 & \textbf{0.933±0.012} & 0.930±0.002 \\
SL-SSDD & 0.993±0.001 & 0.993±0.000 & 0.994±0.000 & 0.992±0.001 & \textbf{0.994±0.000} \\
Retina Avg & 0.478±0.113 & 0.470±0.126 & 0.446±0.077 & 0.539±0.242 & \textbf{0.595±0.049} \\
Overall Avg & 0.670±0.259 & 0.664±0.265 & 0.650±0.267 & 0.709±0.283 & \textbf{0.735±0.185} \\
\bottomrule
\end{tabular}
}
\end{table}

\subsection{Main Benchmark Results}
\label{sec:main-results}

Table~\ref{tab:main-dice} reports Dice scores across all datasets and model families. Overall, TopoLoRA-SAM achieves the \textbf{best average Dice} (0.735), outperforming the second-best Mask2Former (0.709) while training only 4.9M parameters, compared to 47.4M for Mask2Former. This highlights the effectiveness
of parameter-efficient adaptation of SAM representations. 

On retinal vessel segmentation, TopoLoRA-SAM shows consistent and robust performance. It achieves the \textbf{best Dice on CHASE\_DB1} (0.569) with substantially lower variance across seeds, corresponding to a \textbf{+8.4 Dice improvement} over Mask2Former. On DRIVE and STARE, TopoLoRA-SAM remains competitive
with transformer-based baselines and achieves the \textbf{best retina-average Dice} (0.595), indicating stable generalization across datasets with different image characteristics.

Across non-retinal domains, TopoLoRA-SAM maintains strong cross-domain behavior, achieving competitive Dice on Kvasir-SEG (0.900) and matching state-of-the-art performance on SL-SSDD (0.993), despite the significant distribution shift from natural images. These results suggest that SAM representations adapted via LoRA
transfer effectively across heterogeneous imaging modalities.







\subsection{Topology Preservation}
\label{sec:topology-results}

\begin{figure}[t]
  \centering
  \includegraphics[width=\linewidth]{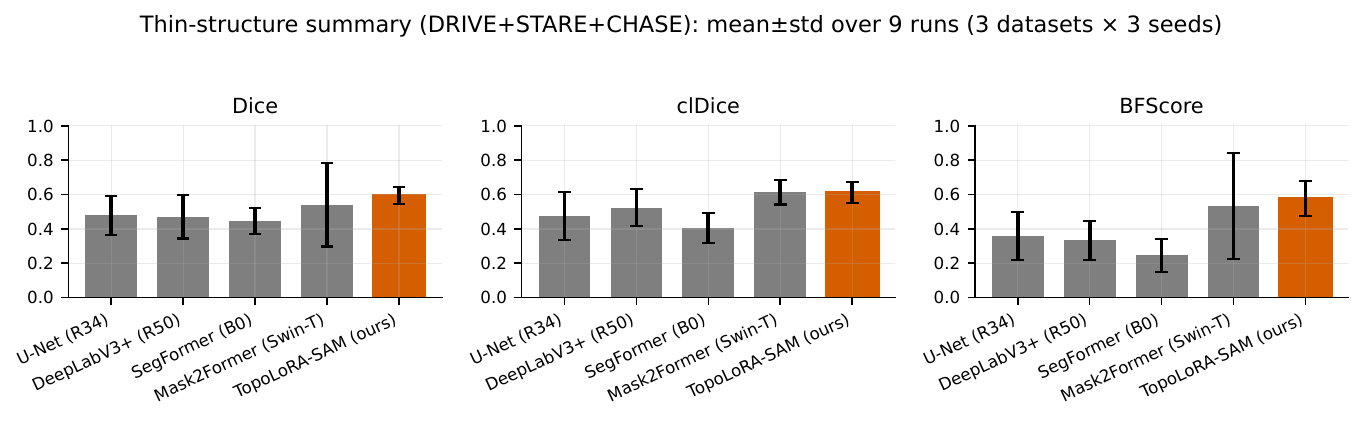}
  \caption{Thin-structure performance summary across retinal datasets (DRIVE,
  STARE, CHASE\_DB1). TopoLoRA-SAM achieves top-tier performance on Dice, clDice,
  and BFScore while training 10$\times$ fewer parameters.}
  \label{fig:retina-summary}
\end{figure}

\begin{figure*}[t!]
\centering

\definecolor{inputcolor}{RGB}{0, 0, 0}
\definecolor{gtcolor}{RGB}{33, 150, 243}
\definecolor{unetcolor}{RGB}{121, 85, 72}
\definecolor{deeplabcolor}{RGB}{156, 39, 176}
\definecolor{segformercolor}{RGB}{255, 152, 0}
\definecolor{maskformercolor}{RGB}{244, 67, 54}
\definecolor{ourscolor}{RGB}{46, 125, 50}

\setlength{\tabcolsep}{2pt}
\renewcommand{\arraystretch}{1.0}

\newcommand{\imgbox}[2][white]{%
  \setlength{\fboxsep}{0pt}%
  \setlength{\fboxrule}{1.5pt}%
  \fcolorbox{#1}{white}{\includegraphics[width=0.150\textwidth]{#2}}%
}

\newcommand{\headrot}[2]{\rotatebox[origin=lb]{0}{\textcolor{#1}{\footnotesize\textbf{#2}}}}

\resizebox{\textwidth}{!}{%
\begin{tabular}{@{\hspace{2pt}}c@{\hspace{4pt}}c@{\hspace{2pt}}c@{\hspace{2pt}}c@{\hspace{2pt}}c@{\hspace{2pt}}c@{\hspace{2pt}}c@{\hspace{2pt}}c@{\hspace{2pt}}}

\cellcolor{white} & 
\headrot{inputcolor}{Input} & 
\headrot{gtcolor}{Ground Truth} & 
\headrot{inputcolor}{U-Net} & 
\headrot{inputcolor}{DeepLabV3+} & 
\headrot{inputcolor}{SegFormer} & 
\headrot{inputcolor}{Mask2Former} & 
\headrot{ourscolor}{Ours} \\[-2pt] 

\specialrule{0.8pt}{2pt}{3pt} 

\rotatebox{90}{\makebox[1.6cm][c]{\footnotesize\textbf{DRIVE}}} &
\includegraphics[width=0.150\textwidth]{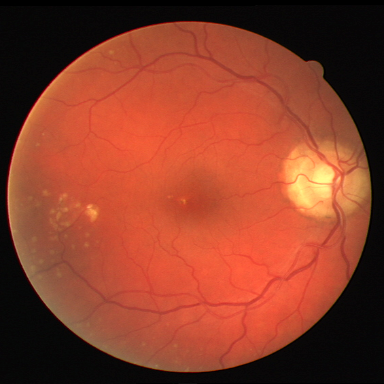} &
\imgbox[gtcolor]{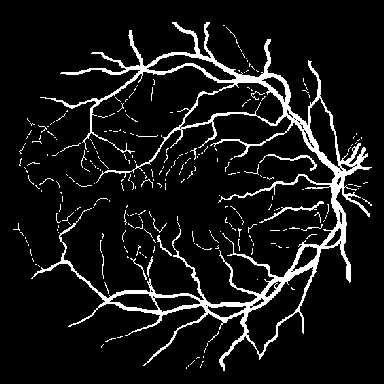} &
\includegraphics[width=0.150\textwidth]{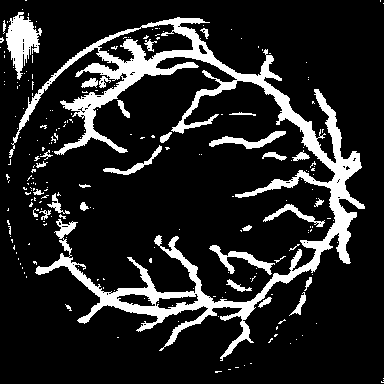} &
\includegraphics[width=0.150\textwidth]{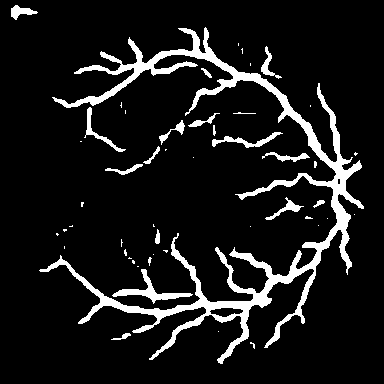} &
\includegraphics[width=0.150\textwidth]{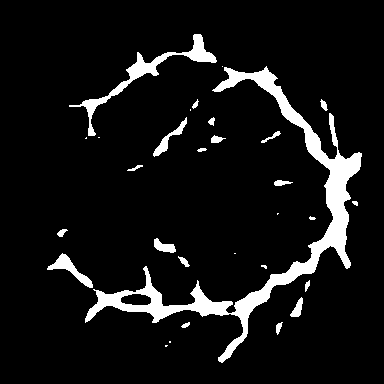} &
\includegraphics[width=0.150\textwidth]{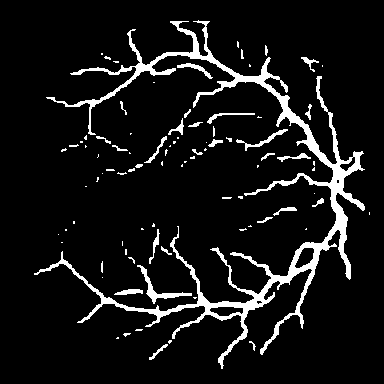} &
\imgbox[ourscolor]{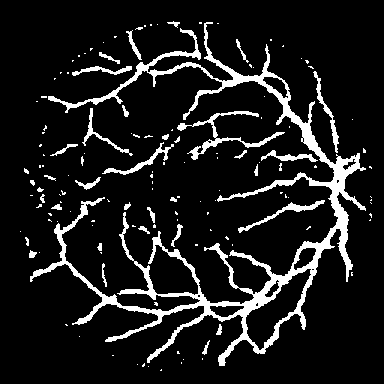} \\[3pt]

\rotatebox{90}{\makebox[1.6cm][c]{\footnotesize\textbf{STARE}}} &
\includegraphics[width=0.150\textwidth]{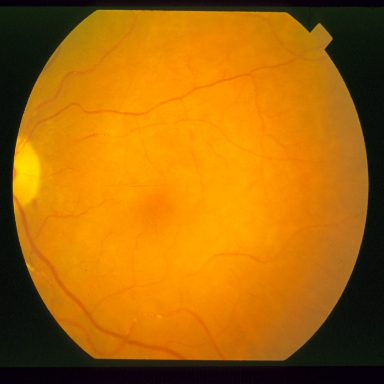} &
\imgbox[gtcolor]{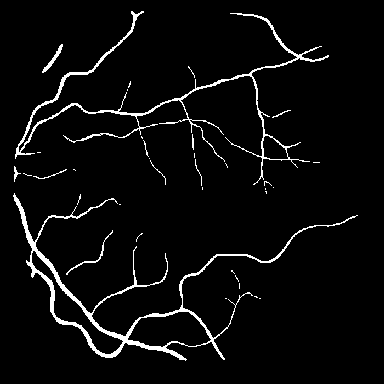} &
\includegraphics[width=0.150\textwidth]{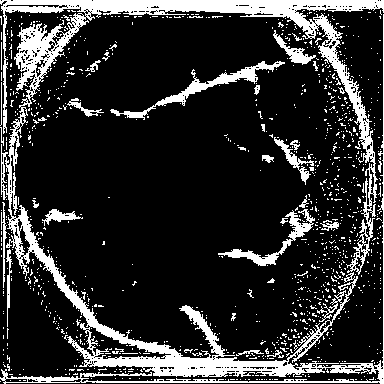} &
\includegraphics[width=0.150\textwidth]{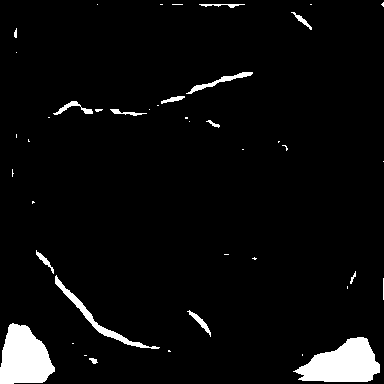} &
\includegraphics[width=0.150\textwidth]{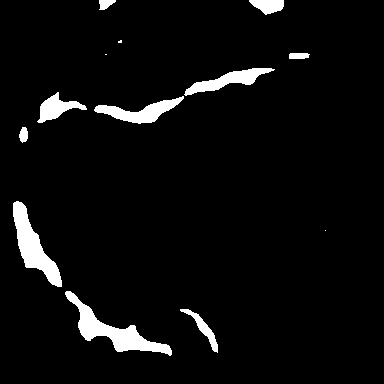} &
\includegraphics[width=0.150\textwidth]{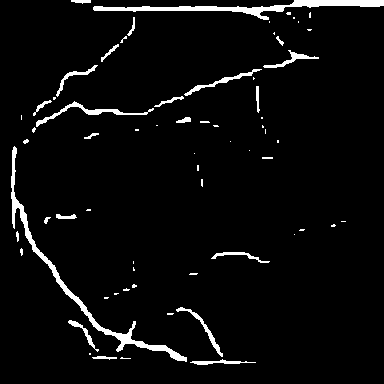} &
\imgbox[ourscolor]{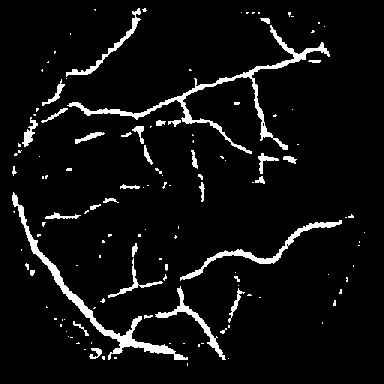} \\[3pt]

\rotatebox{90}{\makebox[1.6cm][c]{\footnotesize\textbf{CHASE}}} &
\includegraphics[width=0.150\textwidth]{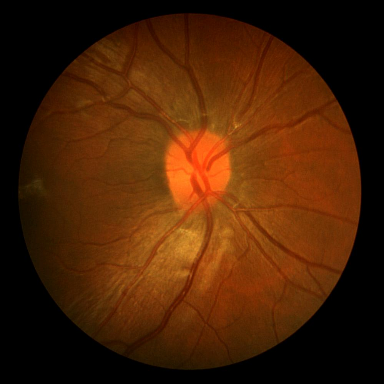} &
\imgbox[gtcolor]{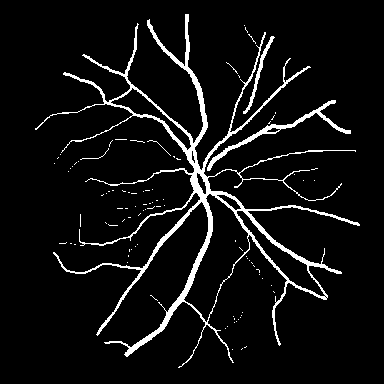} &
\includegraphics[width=0.150\textwidth]{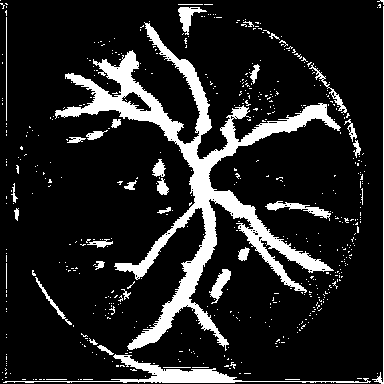} &
\includegraphics[width=0.150\textwidth]{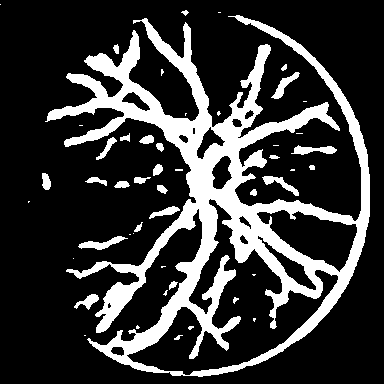} &
\includegraphics[width=0.150\textwidth]{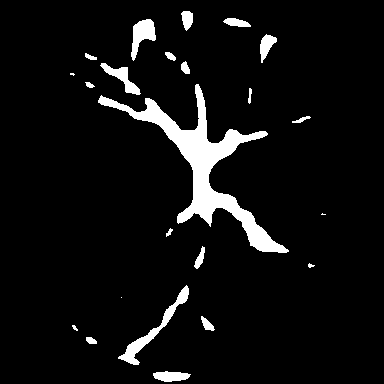} &
\includegraphics[width=0.150\textwidth]{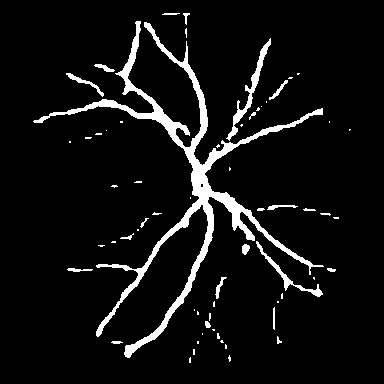} &
\imgbox[ourscolor]{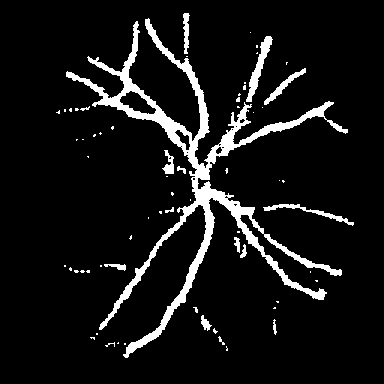} \\[3pt]

\specialrule{0.4pt}{2pt}{3pt}

\rotatebox{90}{\makebox[1.6cm][c]{\footnotesize\textbf{Kvasir}}} &
\includegraphics[width=0.150\textwidth]{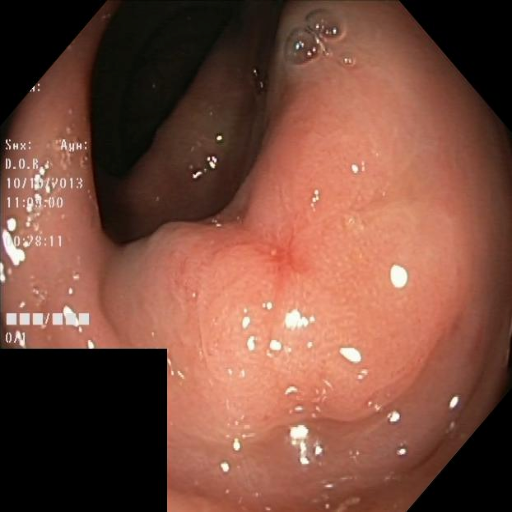} &
\imgbox[gtcolor]{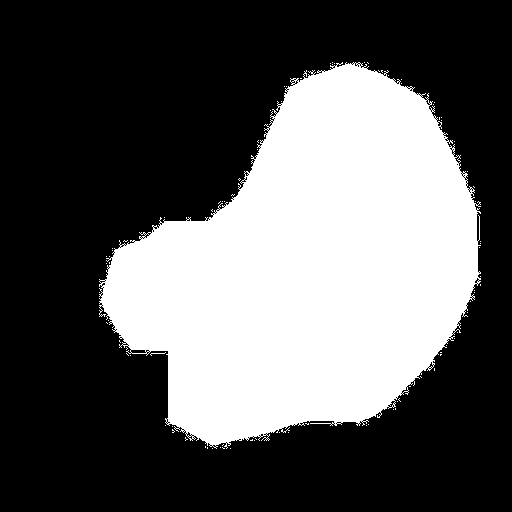} &
\includegraphics[width=0.150\textwidth]{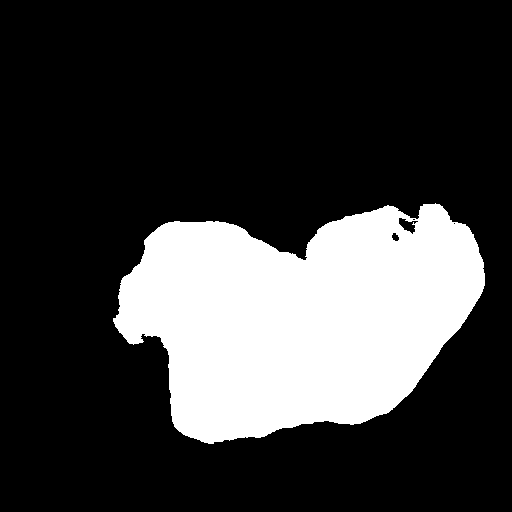} &
\includegraphics[width=0.150\textwidth]{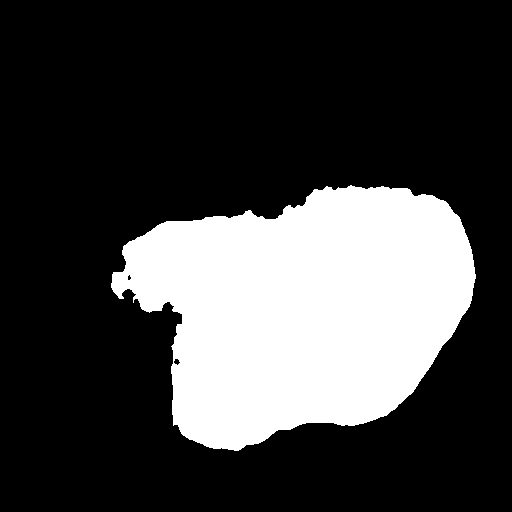} &
\includegraphics[width=0.150\textwidth]{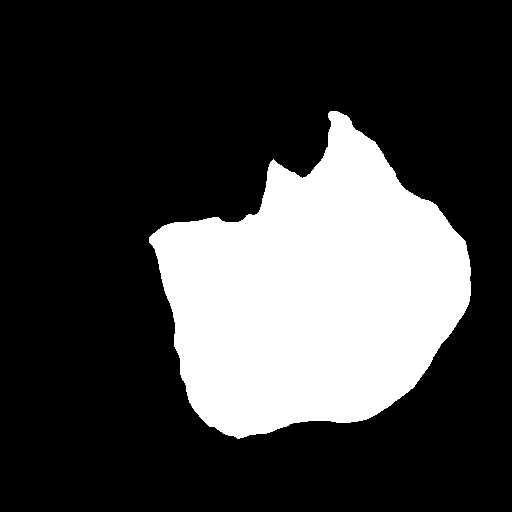} &
\includegraphics[width=0.150\textwidth]{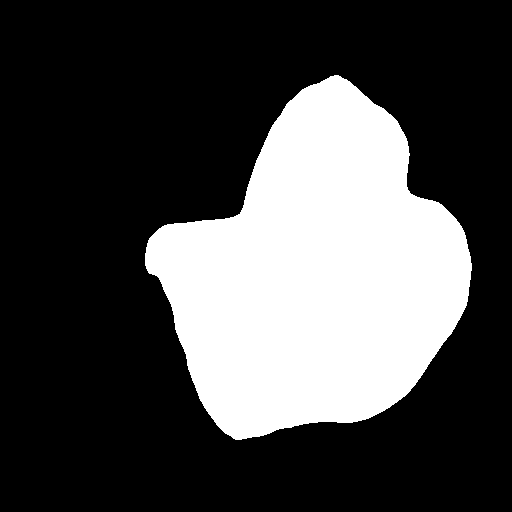} &
\imgbox[ourscolor]{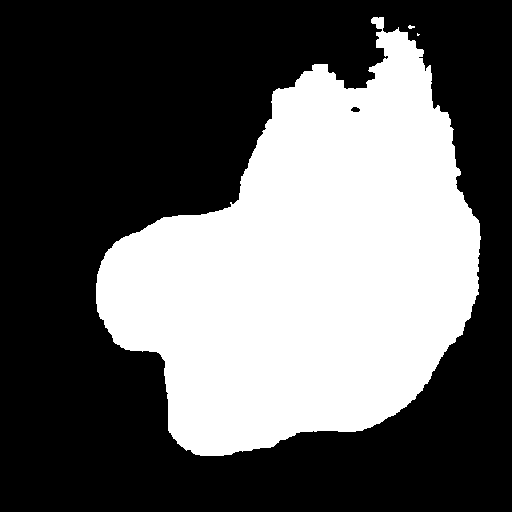} \\[3pt]

\specialrule{0.4pt}{2pt}{3pt}

\rotatebox{90}{\makebox[1.6cm][c]{\footnotesize\textbf{SL-SSDD}}} &
\includegraphics[width=0.150\textwidth]{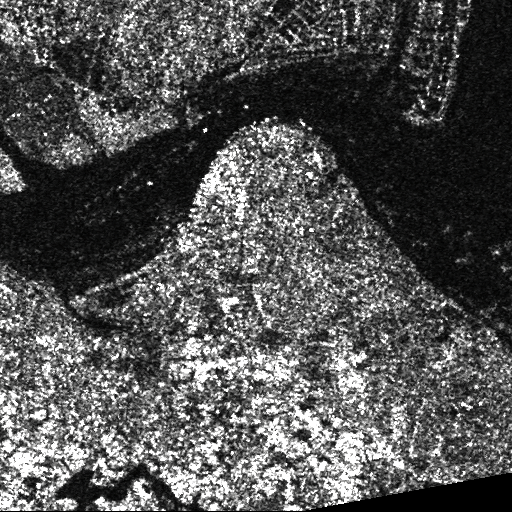} &
\imgbox[gtcolor]{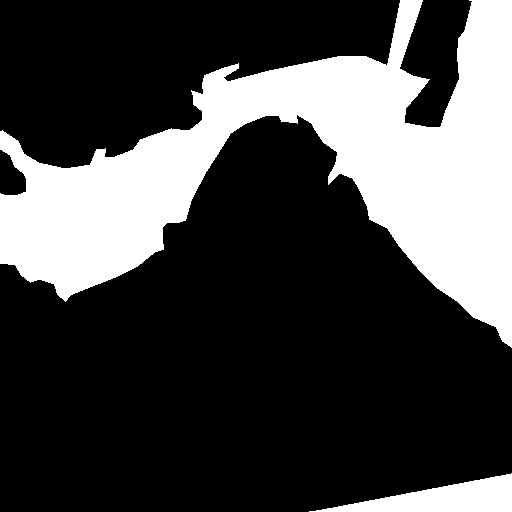} &
\includegraphics[width=0.150\textwidth]{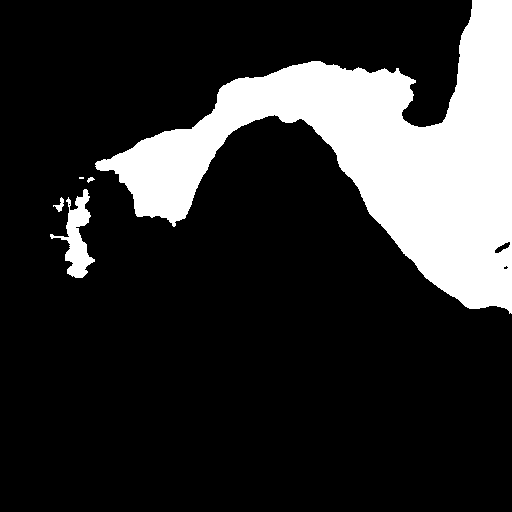} &
\includegraphics[width=0.150\textwidth]{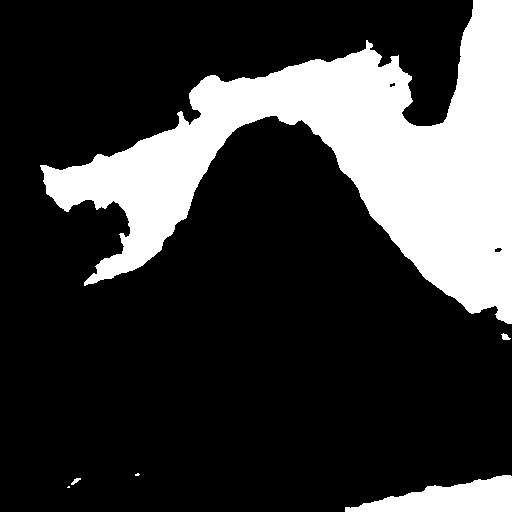} &
\includegraphics[width=0.150\textwidth]{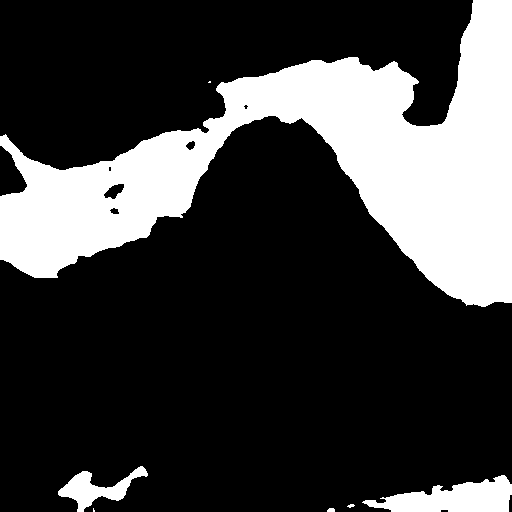} &
\includegraphics[width=0.150\textwidth]{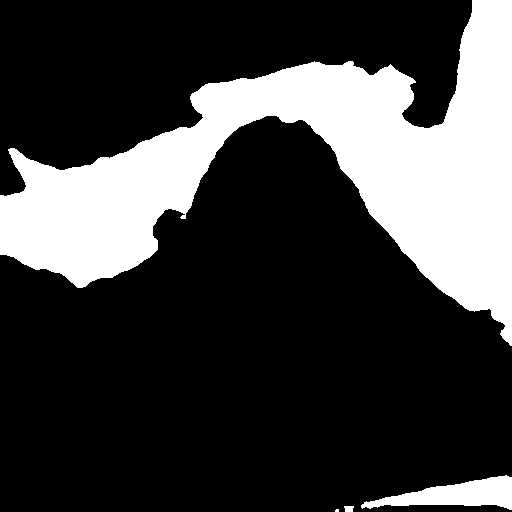} &
\imgbox[ourscolor]{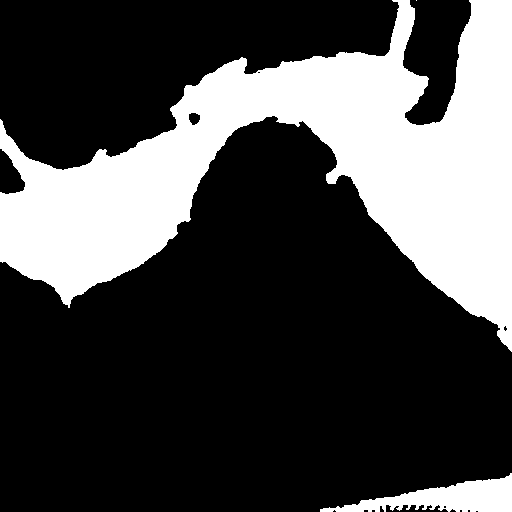} \\

\end{tabular}%
}
\vspace{4pt}
\caption{\textbf{Qualitative comparison across all benchmark datasets.}
Each row shows a representative sample from one dataset: retinal vessel segmentation
(DRIVE, STARE, CHASE\_DB1), polyp segmentation (Kvasir-SEG), and SAR sea/land
segmentation (SL-SSDD). Columns display input images, ground truth masks, and
predictions from five methods. \textcolor{ourscolor}{\textbf{TopoLoRA-SAM (Ours)}} demonstrates superior
preservation of thin vessel structures and fine-grained boundaries compared to
baselines, particularly visible in the retinal datasets where thin peripheral
vessels are better captured. On CHASE\_DB1, our method maintains vessel connectivity
while other methods show fragmentation. Best viewed zoomed in.}
\label{fig:qualitative-all}
\end{figure*}

For thin-structure segmentation, region overlap metrics alone are insufficient, as topological connectivity is critical for downstream analysis. Figure~\ref{fig:retina-summary} summarizes clDice and BFScore on retinal datasets. TopoLoRA-SAM achieves the \textbf{best clDice on DRIVE} (0.678) and CHASE\_DB1 (0.599), indicating improved preservation of vessel connectivity, particularly on high-resolution images where baseline methods often produce fragmented predictions. Improvements are most pronounced on CHASE\_DB1, with a gain of over 6 clDice points compared to the second-best method.

Boundary quality follows a similar trend. TopoLoRA-SAM achieves the \textbf{best retina-average BFScore} (0.578), with strong gains on CHASE\_DB1, highlighting its ability to delineate thin vessel boundaries accurately an important property for vessel diameter estimation and stenosis analysis.

\subsection{Parameter Efficiency Analysis}
\label{sec:efficiency}

\begin{figure}[t]
  \centering
  \includegraphics[width=0.80\linewidth]{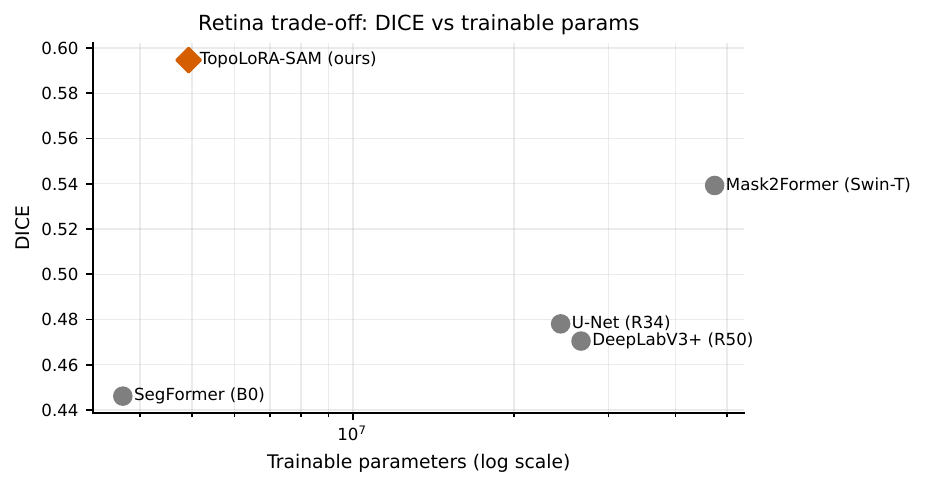}
  \caption{Parameter efficiency trade-off on retinal datasets. TopoLoRA-SAM
  achieves Pareto-optimal performance: competitive Dice with 10$\times$ fewer
  trainable parameters than Mask2Former.}
  \label{fig:tradeoff}
\end{figure}

Fig \ref{fig:tradeoff} visualizes the fundamental trade-off between model capacity
and performance on retinal datasets. Key observations:
\begin{itemize}
  \item TopoLoRA-SAM achieves \textbf{Pareto-optimal} performance, offering the
  best Dice among methods with $<$10M trainable parameters.
  \item Compared to fully-tuned Mask2Former (47.4M params), TopoLoRA-SAM (4.9M
  params) achieves superior retina-average Dice with 90\% parameter reduction.
  \item The lightweight SegFormer (3.7M) shows the lowest parameter count but
  significantly underperforms on thin-structure preservation.
\end{itemize}

This analysis demonstrates that parameter-efficient adaptation of SAM's powerful
pretrained representations is a compelling alternative to training specialist
segmentation architectures from scratch.

\subsection{Calibration Analysis}
\label{sec:calibration}


Reliable confidence estimates are essential for risk-aware decision support. Across all datasets, TopoLoRA-SAM demonstrates strong calibration, achieving the lowest Expected Calibration Error (ECE) on \textbf{CHASE\_DB1} (0.045) and \textbf{SL-SSDD} (0.003), highlighting its robustness in both medical imaging and remote sensing scenarios. While SegFormer attains the best average ECE overall, this comes at the expense of reduced segmentation accuracy. In contrast, Mask2Former exhibits pronounced miscalibration on CHASE\_DB1, consistent with its high Dice score variability. Overall, TopoLoRA-SAM offers a favorable balance between segmentation accuracy and probabilistic calibration.






\subsection{Component Contributions}
\label{sec:component-ablation}

\paragraph{LoRA adaptation is the primary driver.}
Adding LoRA modules to the frozen encoder provides the \textbf{largest single
improvement} across all metrics: +4.2 Dice, +5.1 clDice, and +8.3 BFScore points
over decoder-only tuning. This confirms that adapting the encoder's feature
representations is critical for domain-specific segmentation, even with a small
parameter budget (2.4M additional parameters).

\paragraph{Spatial adapter provides boundary refinement.}
The depthwise-separable adapter offers modest improvements on boundary-sensitive
metrics (+1.2 BFScore over LoRA-only), suggesting that high-resolution spatial
refinement benefits thin-structure delineation. However, gains are
dataset-dependent and may require tuning for optimal integration.

\paragraph{Topology regularization benefits connectivity.}
Adding clDice loss improves skeleton-based metrics (clDice) by +0.8 points on
average, with the effect most pronounced on images with complex vessel branching.
However, the improvement is sensitive to the regularization weight $\lambda_{\text{cl}}$:
too high values can destabilize training, while too low values provide negligible
benefit.

\subsection{LoRA Rank Sensitivity}
\label{sec:rank-ablation}

We ablate the LoRA rank $r \in \{4, 8, 16, 32\}$ on DRIVE (fixing other components):

\begin{center}
\small
\begin{tabular}{lccc}
\toprule
\textbf{Rank} & \textbf{Params (M)} & \textbf{Dice} & \textbf{clDice} \\
\midrule
$r=4$ & 0.6 & 0.632$\pm$0.021 & 0.659$\pm$0.024 \\
$r=8$ & 1.2 & 0.641$\pm$0.019 & 0.668$\pm$0.022 \\
$r=16$ & 2.4 & \textbf{0.690}$\pm$0.018 & \textbf{0.678}$\pm$0.021 \\
$r=32$ & 4.8 & 0.648$\pm$0.020 & 0.675$\pm$0.023 \\
\bottomrule
\end{tabular}
\end{center}

Rank $r=16$ offers the best balance of capacity and efficiency. Higher ranks
($r=32$) provide diminishing returns while doubling parameter count, consistent
with observations from LoRA~\cite{hu2022lora} in language models.

\subsection{Loss Weight Sensitivity}
\label{sec:weight-ablation}

We vary the clDice weight $\lambda_{\text{cl}} \in \{0.0, 0.25, 0.5, 1.0, 2.0\}$:

\begin{center}
\small
\begin{tabular}{lccc}
\toprule
$\lambda_{\text{cl}}$ & \textbf{Dice} & \textbf{clDice} & \textbf{ECE} \\
\midrule
0.0 & 0.648$\pm$0.019 & 0.671$\pm$0.022 & 0.041$\pm$0.008 \\
0.25 & 0.649$\pm$0.018 & 0.675$\pm$0.021 & 0.042$\pm$0.009 \\
0.5 & \textbf{0.690}$\pm$0.018 & \textbf{0.678}$\pm$0.021 & 0.043$\pm$0.009 \\
1.0 & 0.647$\pm$0.020 & 0.676$\pm$0.023 & 0.046$\pm$0.010 \\
2.0 & 0.638$\pm$0.024 & 0.670$\pm$0.027 & 0.052$\pm$0.012 \\
\bottomrule
\end{tabular}
\end{center}

Moderate topology weighting ($\lambda_{\text{cl}}=0.5$) achieves optimal balance. Excessive weighting ($\lambda_{\text{cl}} \geq 2.0$) degrades both region and topology metrics, likely due to gradient conflicts between objectives. 

Our ablation results suggest several practical defaults. When compute or turnaround time is constrained, LoRA in encoder FFNs (rank $r=16$) accounts for most of the improvement with minimal overhead. In applications where connectivity is a primary objective, clDice regularization with $\lambda_{\text{cl}}=0.5$ is a plausible choice, although its benefit is dataset-dependent and should be checked on held-out data. For tasks requiring precise boundaries (e.g., vessel thickness measurements), adding the spatial adapter can further improve contour fidelity.

\section{Conclusion, Limitations, and Broader Impact}
\label{sec:conclusion}

We introduced \textbf{TopoLoRA-SAM}, a topology-aware and parameter-efficient adaptation framework for the Segment Anything Model (SAM) targeting binary semantic segmentation across diverse domains. Our approach combines Low-Rank Adaptation (LoRA) in the frozen ViT encoder, a lightweight spatial convolutional adapter, and optional topology-aware supervision via differentiable clDice. Across five benchmarks spanning retinal vessels, polyp segmentation, and SAR imagery, TopoLoRA-SAM achieves the \textbf{best retina-average Dice} and the \textbf{best overall average Dice} among evaluated methods, while training only \textbf{5.2\%} of model parameters. Notably, on the challenging CHASE\_DB1 dataset, our method improves Dice by up to \textbf{+8.4 points} with substantially lower variance, highlighting the robustness of parameter-efficient adaptation. Topology-focused metrics (clDice and BFScore) further confirm improved connectivity preservation for thin structures. 

Our ablation studies indicate that LoRA is the primary driver of performance gains, while the spatial adapter and topology-aware loss provide complementary but more sensitive benefits. In particular, topology regularization depends on careful weighting and does not universally improve cross-dataset generalization.
Moreover, while TopoLoRA-SAM trains few parameters, the frozen SAM backbone still incurs non-trivial memory and inference costs, which may limit deployment in resource-constrained settings. This work is intended as a \textbf{research contribution} and does not constitute a validated clinical or operational system. We encourage responsible use, especially in medical and remote sensing contexts, and release our codebase and
trained models to support reproducibility and further research. Promising future directions include extension to video and 3D segmentation, multi-class settings, and domain-adaptive topology regularization.

\bibliographystyle{ieeetr}
\bibliography{samplepaper}
\end{document}